\documentclass[conference]{IEEEtran}
\usepackage{cite}
\usepackage{amsmath,amssymb,amsfonts}
\usepackage{algorithmic}
\usepackage{graphicx}
\usepackage{textcomp}
\usepackage{xcolor}
\def\BibTeX{{\rm B\kern-.05em{\sc i\kern-.025em b}\kern-.08em
    T\kern-.1667em\lower.7ex\hbox{E}\kern-.125emX}}

\usepackage[ruled,vlined]{algorithm2e}
\usepackage{booktabs}
\usepackage{multirow}
\usepackage{lscape}
\usepackage{pifont}

\usepackage[normalem]{ulem}
\usepackage{subcaption}
\usepackage[table]{colortbl}

\usepackage{todonotes}
    
\begin{document}

\title{
Federated Multimodal Learning with Dual Adapters and Selective Pruning for Communication and Computational Efficiency
}


\author{
\IEEEauthorblockN{
Duy Phuong Nguyen\IEEEauthorrefmark{1}, 
J. Pablo Muñoz \IEEEauthorrefmark{2},
Tanya Roosta\IEEEauthorrefmark{3} and
Ali Jannesari\IEEEauthorrefmark{1}
}

\IEEEauthorblockA{
\IEEEauthorrefmark{1} Iowa State University, 
\IEEEauthorrefmark{2} Intel Labs,
\IEEEauthorrefmark{3} Amazon
}

\IEEEauthorblockA{
dphuong@iastate.edu,
pablo.munoz@intel.com,
tanya.roosta@gmail.com,
jannesar@iastate.edu
}
}

\newcommand{\proj}{\textsc{FedDLP}\xspace}

\newcommand{\cmark}{\ding{51}}%
\newcommand{\xmark}{\ding{55}}%

\maketitle

\begin{abstract}

Federated Learning (FL) enables collaborative learning across distributed clients while preserving data privacy. However, FL faces significant challenges when dealing with heterogeneous data distributions, which can lead to suboptimal global models that fail to generalize across diverse clients. In this work, we propose a novel framework designed to tackle these challenges by introducing a dual-adapter approach. The method utilizes a larger local adapter for client-specific personalization and a smaller global adapter to facilitate efficient knowledge sharing across clients. Additionally, we incorporate a pruning mechanism to reduce communication overhead by selectively removing less impactful parameters from the local adapter. Through extensive experiments on a range of vision and language tasks, our method demonstrates superior performance compared to existing approaches. It achieves higher test accuracy, lower performance variance among clients, and improved worst-case performance, all while significantly reducing communication and computation costs. Overall, the proposed method addresses the critical trade-off between model personalization and generalization, offering a scalable solution for real-world FL applications.


\end{abstract}

\section{Introduction}~\label{sec:introduction}

\vspace{-1.0em}

\begin{figure*}[ht]
\begin{center}

\centerline{\includegraphics[width=0.7\textwidth]{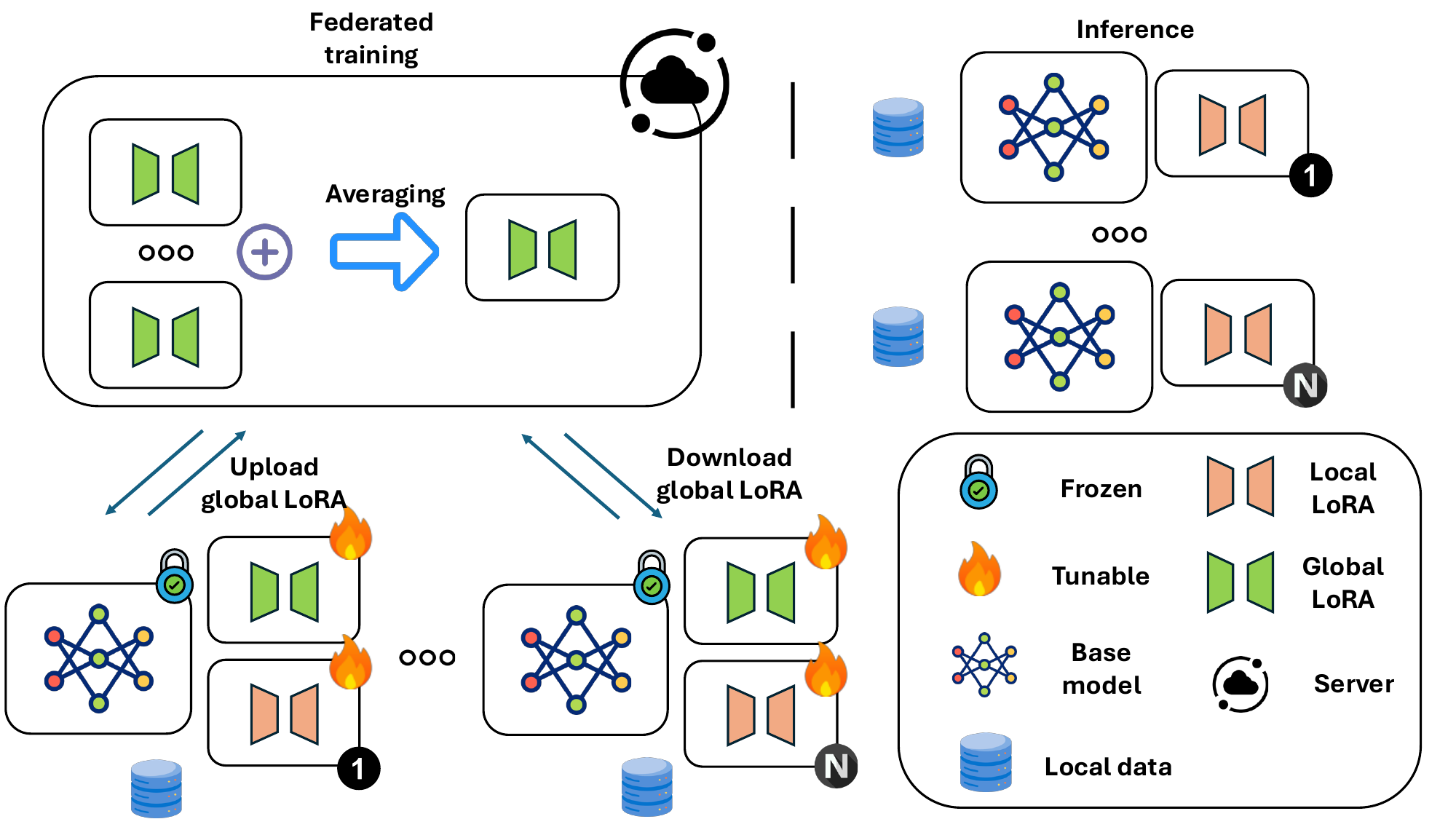}}

\caption{Overview of our proposed method in the federated learning setup. Each client is equipped with a base model consisting of two separate LoRA adapters: a larger, tunable local LoRA adapter (orange) that is kept private to each client and a smaller, tunable global LoRA adapter (green) that is communicated to and from the central server. During training, the global adapter is aggregated across all clients, while the local adapter remains personalized. At inference, the final model is a combination of the frozen base model and personalized local LoRA for each client. The figure is best viewed in color.
}

\label{fig:overview}

\vspace{-3.0em}

\end{center}
\end{figure*}

Federated Learning (FL)\cite{mcmahan2017fedavg} has emerged as a powerful paradigm for enabling distributed clients to collaboratively train models while preserving data privacy. Typically, the global model is updated by aggregating client-specific updates, but this approach can be resource-intensive, particularly when applied to large models such as CLIP\cite{radford2021clip}, a state-of-the-art model for vision and language tasks. These challenges are further exacerbated by the non-IID~(non-Independent and Identically Distributed) nature of client data, which can lead to significant discrepancies in local model updates and degrade global model performance~\cite{mcmahan2017fedavg, li2020fedprox}. To address these issues, personalized Federated Learning (pFL) has been proposed~\cite{tan2023pfl}.

Traditional FL approaches~\cite{mcmahan2017fedavg, li2020convergence_noniid} aim to train a single global model evaluated on a common test set, assuming that all clients share similar data distributions. However, in real-world scenarios, client data is often non-IID, making a single global model inadequate. In pFL, the objective is to learn a unique model for each client, tailored to their local data distribution~\cite{tan2023pfl, yao2024rethinking}, ensuring personalized performance. Unlike traditional FL, pFL evaluates models on each client’s own test set to better align with individual data distributions.

Our proposed method, \textbf{Fed}erated \textbf{D}ual \textbf{L}oRA \textbf{P}runing (\proj), operates within the pFL framework. By tailoring the model to each client’s local data distribution, \proj addresses the limitations of homogeneous global models. Specifically, \proj focuses on fine-tuning LoRA adapters attached to the CLIP model while integrating a pruning mechanism to enhance communication efficiency and computational scalability.


Low-rank Adaptation (LoRA)\cite{hu2022lora} has proven to be an effective method for reducing the number of parameters updated during model fine-tuning by leveraging low-rank matrices. This significantly reduces communication overhead in FL, as only low-rank updates, rather than full model parameters, are transmitted between clients and the server. However, applying homogeneous LoRA adapters to all clients in FL fails to account for the non-IID nature of client data. This often results in overfitting to local data and poor generalization across clients\cite{nguyen2024flora}. The issue arises because clients typically access only a subset of the overall data distribution, causing homogeneous LoRA layers to overfit to specific patterns unrepresentative of the global distribution~\cite{ li2020convergence_noniid, cho2023heterolora}.


To mitigate these issues, we draw inspiration from SoRA (Sparse Low-rank Adaptation)~\cite{ding2023sora}, which prunes LoRA layers based on their importance. In \proj, we adopt a similar approach, pruning local LoRA layers at each client to focus on the most critical components for the client’s specific data. This reduces communication overhead by transmitting only the essential parts of the local adapter back to the server. However, the client-specific pruning introduces heterogeneity in the LoRA structures, complicating aggregation at the server.



To address this, we take inspiration from FedDAT~\cite{chen2024feddat}, which fine-tunes foundation models in FL using adapters. While FedDAT uses homogeneous adapters, we adapt their strategy to handle pruned and heterogeneous LoRA adapters. In \proj, each client fine-tunes a larger local LoRA adapter for personalization and maintains a smaller global LoRA adapter for knowledge sharing. Only the global adapter is aggregated at the server, capturing generalized knowledge while allowing the local adapter to retain client-specific updates. An overview of this approach is shown in Figure~\ref{fig:overview}.


By combining LoRA adapters with pruning, \proj significantly reduces communication costs and enhances computational efficiency, all while maintaining strong performance across clients with diverse, non-IID data. The framework also addresses the challenges of aggregating pruned, heterogeneous adapters, enabling the system to balance local personalization and global generalization effectively.



In summary, the contributions of our work are as follows: 

\begin{itemize} 
    \item We propose \proj, a novel FL framework that integrates LoRA adapters with a pruning mechanism to address communication efficiency and computational complexity, particularly for large models such as CLIP. 
    
    \item We address the challenge of aggregating pruned LoRA adapters from clients with non-IID data by maintaining a separate global adapter that can be consistently aggregated across clients, while allowing local adapters to be fine-tuned and pruned using local data. 
    
    \item We provide extensive experimental results demonstrating that \proj reduces communication overhead and computational costs while maintaining competitive performance in diverse FL settings. \end{itemize}

\section{Related Work}~\label{sec:related}

\vspace{-1.5em}

\subsection{Federated Learning (FL)}

Federated Learning (FL) is a distributed machine learning paradigm that enables multiple clients to collaboratively train a shared model while keeping their data localized. The key idea is to perform local model training on each client's data and then aggregate updates on a central server. This approach preserves data privacy and security, as raw data remains on the clients' devices. To improve the communication efficiency of FL, Konečný et al.~\cite{konečný2017flstrategiesimproving} introduced strategies such as model quantization and sparsification~\cite{li2020convergence_noniid}. Federated averaging (FedAvg)~\cite{mcmahan2017fedavg}, a foundational method, further enhances efficiency and scalability by averaging model updates across clients.

Several FL methods have been developed to address challenges in efficiency and heterogeneity. For example, FedRep~\cite{collins2021fedrep} separates local and global representations, while FedRoD~\cite{chen2022fedrod} focuses on parameter sharing. However, \proj improves upon these approaches by incorporating structured pruning to reduce communication costs while maintaining strong performance. Similarly, FedAvgM~\cite{hsu2019fedavgm} and FedProx~\cite{li2020fedprox} address data heterogeneity through weighted averaging or regularization but lack mechanisms for aggregating pruned LoRA adapters. Lastly, FLoRA~\cite{nguyen2024flora} introduces LoRA adapters in FL but does not include pruning or address the complexities of adapter aggregation in heterogeneous settings. By contrast, \proj’s dual-adapter design effectively balances personalization and generalization while addressing these limitations.

\vspace{-1.5em}

\subsection{Vision and Language Models (VLMs)}

\vspace{-0.em}

Multi-modal learning, which integrates information from diverse data types such as text, images, and audio, has transformed artificial intelligence~\cite{vinyals2015pointer,xu2015show}. By mimicking the human ability to process multiple sensory inputs simultaneously, multi-modal systems enable a more comprehensive understanding of the world~\cite{antol2015vqa,li2021albef,kim2021vilt}. This synergy between modalities results in richer representations and more robust learning, driving breakthroughs in tasks that demand nuanced understanding of both context and content~\cite{lu2019vilbert,tan2019lxmert}.

A prominent example of multi-modal learning is the Contrastive Language–Image Pre-training (CLIP) model developed by OpenAI~\cite{radford2021clip}. CLIP exemplifies the power of multi-modality by combining vision and language to achieve state-of-the-art performance. Its architecture consists of two encoders—an image encoder and a text encoder—trained jointly to map visual and textual data into a shared embedding space. This shared space enables CLIP to understand and associate visual concepts with natural language descriptions, making it remarkably versatile for a wide range of vision and language tasks.



\vspace{-0.4em}

\subsection{Low-Rank Adaptation (LoRA)}

Low-Rank Adaptation (LoRA) is a parameter-efficient fine-tuning technique that adapts pre-trained models by introducing low-rank matrices while freezing the original model parameters. This approach significantly reduces the number of parameters that need to be fine-tuned, lowering both computational and memory requirements. Hu et al.~\cite{hu2022lora} demonstrated that LoRA achieves competitive performance with far fewer parameters compared to traditional fine-tuning methods. By focusing on low-rank updates, LoRA enables efficient model adaptation in resource-constrained environments, making it particularly well-suited for FL scenarios where communication bandwidth is limited~\cite{wu2024fedlora,yi2024pfedlora}.

The integration of LoRA and other Parameter-Efficient Fine-Tuning (PEFT) techniques for VLMs in FL has gained significant attention~\cite{wu2024fedlora,yi2024pfedlora}. Recent studies have shown that incorporating LoRA in FL settings can optimize both communication and computational efficiency~\cite{sun2024improving}. For example, FedPETuning~\cite{zhang2023fedpetuning}, an FL method leveraging LoRA, demonstrated improved model accuracy while reducing communication costs. Additionally, research on PEFT techniques in FL has highlighted their effectiveness in adapting large models across distributed environments. Studies such as those by Bai et al.~\cite{bai2024federatedfinetuningllm,babakniya2023slora} and others emphasize the potential of LoRA to minimize parameter size and enhance the efficiency of training and communication in FL systems. These advancements are pivotal for scaling FL to support large-scale models effectively.

\subsection{Model Pruning}

Model pruning is a technique for reducing the size of neural networks by removing less important weights or neurons, resulting in more efficient models without a significant loss in performance. Pruning can be performed during or after training, using criteria such as weight magnitude or gradient-based importance to identify which parts of the model to remove. Renda et al.~\cite{renda2020comparingrewindingfinetuningneural} investigated various pruning strategies, comparing the effectiveness of weight rewinding and fine-tuning in maintaining model accuracy. Block pruning, where entire blocks of weights are removed, has been shown to effectively reduce model complexity while preserving performance~\cite{men2024shortgpt, zhong2024blockpruner}.

Several advanced pruning techniques have been developed to optimize model efficiency further. For example, Wanda~\cite{sun2024wanda} calculates the importance score of each weight by combining its magnitude with the norm of the associated input activation, enabling the selective removal of less critical weights. SparseGPT~\cite{frantar2023sparsegpt} and LoRAPrune~\cite{zhang2023loraprune} focus on unstructured pruning for large language models, eliminating parameters independently of the model's internal structure. Notably, SoRA~\cite{ding2023sora} extends pruning to LoRA layers, allowing them to be selectively pruned while fine-tuning on downstream tasks, demonstrating the potential for efficient adaptation in resource-constrained settings.


\section{Motivation}~\label{sec:motivation}

\vspace{-1.0em}

\begin{figure}[ht]
\begin{center}

\vspace{-2.0em}

\centerline{\includegraphics[width=0.35\textwidth]{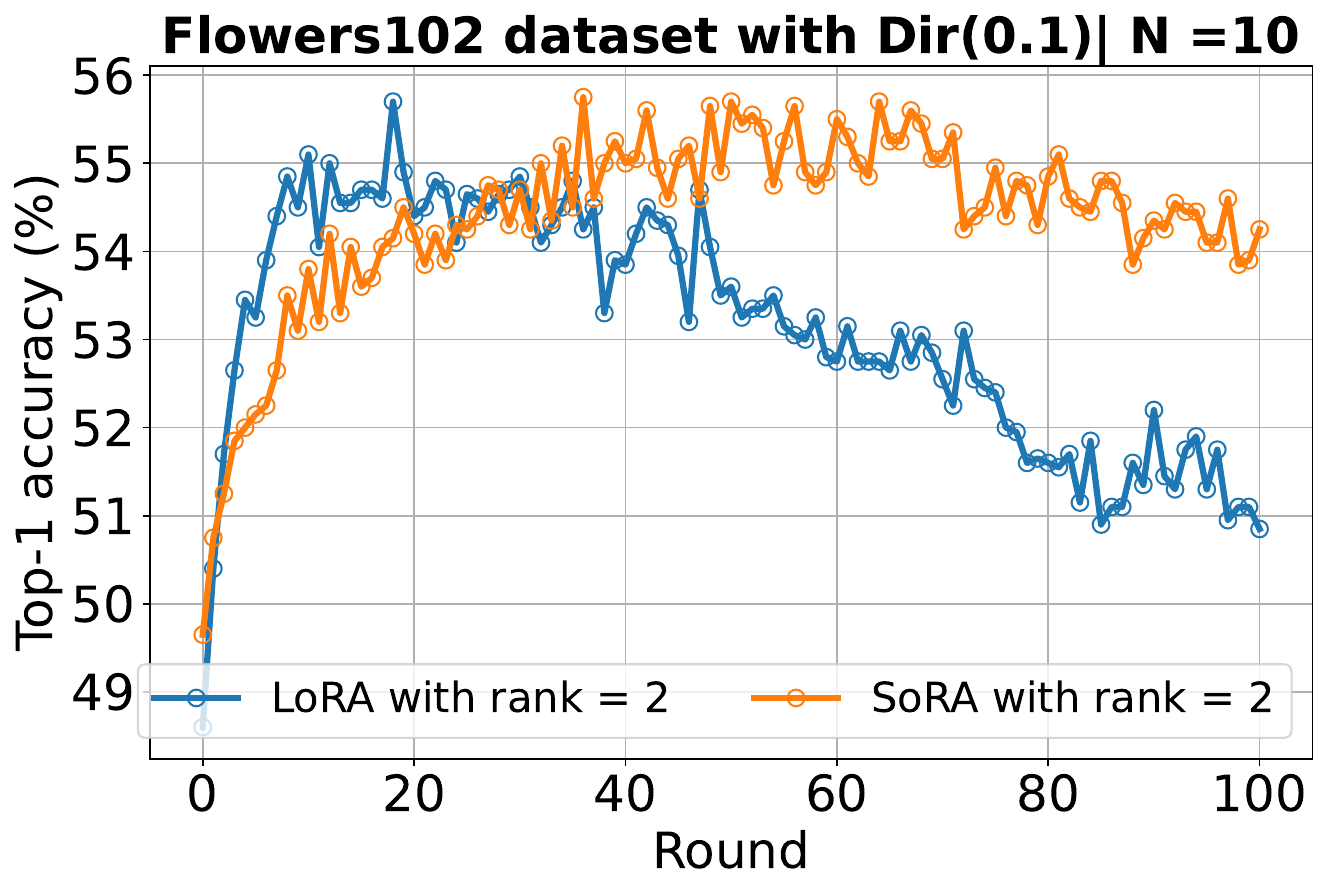}}

\caption{Comparison of vanilla LoRA and SoRA (both with rank = 2) in pFL on the Flowers102 dataset~\cite{nilsback2008flowers}. Both methods are trained locally without communication. While LoRA initially improves, it suffers from overfitting and performance degradation as training progresses, especially on non-IID data. In contrast, SoRA, with its sparsity mechanism, demonstrates better generalization and maintains higher Top-1 accuracy across rounds, highlighting the effectiveness of structured pruning in handling heterogeneous client data. Both adapeters are applied to the image encoder of CLIP~\cite{radford2021clip}.}

\vspace{-3.0em}

\label{fig:motivation}
\end{center}
\end{figure}

In personalized Federated Learning (pFL), a key challenge is training models that can generalize effectively across clients with non-IID data while ensuring that each client benefits from personalized updates~\cite{mcmahan2017fedavg, tan2023pfl}. Parameter-efficient techniques, such as LoRA~\cite{hu2022lora}, have been explored to reduce the number of trainable parameters in large models like CLIP, making them suitable for FL settings where communication and computation resources are constrained.

To evaluate the effectiveness of LoRA in pFL, we conducted a preliminary study using vanilla LoRA with a rank of 2. The Flowers102 dataset~\cite{nilsback2008flowers} was partitioned into 10 clients using a Dirichlet distribution with $\beta=0.1$ to simulate a non-IID data setting. LoRA was applied to the image encoder of the CLIP model, and training was performed locally at each client without any communication or parameter sharing. Each client trained its model exclusively on its local data.

As shown in Figure~\ref{fig:motivation}, LoRA demonstrated initial improvements in Top-1 accuracy on the Flowers102 dataset but suffered from performance degradation in later rounds. This indicates that, in a non-communication scenario, LoRA overfits to local data and struggles to adapt to the heterogeneous distributions across clients.

We next applied SoRA~\cite{ding2023sora}, which incorporates a structured pruning mechanism, in the same setup with the same rank. As illustrated in Figure~\ref{fig:motivation}, SoRA consistently outperformed vanilla LoRA by achieving higher Top-1 accuracy, particularly in later training rounds. The improvement can be attributed to SoRA's sparsity mechanism, which focuses on pruning less important parts of the model, allowing it to generalize better to the local data distribution while mitigating overfitting.

This comparison highlights the limitations of using homogeneous LoRA adapters in pFL. While LoRA reduces the number of trainable parameters, its inability to handle data heterogeneity results in performance degradation. In contrast, SoRA’s sparsity mechanism enables better generalization and demonstrates the value of structured pruning in pFL settings.

However, the use of client-specific pruning in SoRA introduces new challenges in aggregating pruned LoRA adapters at the central server. Since each client prunes different parts of the model based on its local data, straightforward aggregation of the pruned adapters leads to performance degradation due to structural inconsistency across clients.

Motivated by the performance gap between LoRA and SoRA and inspired by FedDAT~\cite{chen2024feddat}, we propose \proj, a framework that addresses these challenges by combining structured pruning with a dual-adapter mechanism. In \proj, each client uses a larger local LoRA adapter for personalization and a smaller global LoRA adapter for knowledge sharing. The local adapter is pruned based on client-specific data, while the global adapter remains consistent across all clients and is aggregated at the server. This design balances local adaptation with global generalization, improving communication efficiency and model performance in heterogeneous pFL environments.

By introducing this dual-adapter mechanism and addressing the aggregation challenge, \proj advances the state-of-the-art in pFL, making it more scalable and effective for large models with non-IID data distributions.

\section{Preliminaries}~\label{sec:prelim}

\vspace{-2.0em}

\subsection{Zero-shot evaluation}

Following the approach outlined in the CLIP paper~\cite{radford2021clip}, we perform zero-shot evaluation by leveraging the names of all classes in each dataset as potential text pairings. The goal is to predict the most likely (image, text) pair based on the CLIP model’s capabilities. Specifically, the text encoder processes inputs in the format "a photo of a [class]," where [class] represents the category of the image in the dataset. This allows CLIP to align the embedding spaces of images and their corresponding textual descriptions labeled as "[class]," enabling classification without requiring task-specific fine-tuning.


First, the feature embedding of an image \(\mathbf{x}\) is computed using the image encoder \(\mathcal{I}(\cdot)\):

\begin{equation}
\mathbf{e}_\text{image} = \mathcal{I}(\mathbf{x}),
\label{eq:image_embedding}
\end{equation}

For a given class \(i\), let \(\mathbf{t}_i\) represent the word embedding vector of the class: \([\text{class}]_i\), where \(i \in [1, K]\), and \(K\) is the total number of classes. The text embeddings for all class labels are obtained using the text encoder \(\mathcal{T}(\cdot)\):

\begin{equation}
\mathbf{e}_{\text{text}, i} = \mathcal{T}(\mathbf{t}_i) \quad \text{for} \; i \in [1, K].
\label{eq:text_embeddings}
\end{equation}

Next, we compute the cosine similarity between the image embedding \(\mathbf{e}_\text{image}\) and each text embedding \(\mathbf{e}_{\text{text}, i}\), scaling the results by a temperature parameter \(\tau_{CLIP}\), learned during CLIP training:


\begin{equation}
\cos(\mathbf{e}_\text{image}, \mathbf{e}_{\text{text}, i}) = \frac{\mathbf{e}_\text{image} \cdot \mathbf{e}_{\text{text}, i}}{\|\mathbf{e}_\text{image}\|_2 \|\mathbf{e}_{\text{text}, i}\|_2},
\label{eq:cosine_similarity}
\end{equation}

\begin{equation}
s_i = \frac{\cos(\mathbf{e}_\text{image}, \mathbf{e}_{\text{text}, i})}{\tau_{CLIP}},
\label{eq:scaled_similarity}
\end{equation}

The predicted class \(\hat{y}\) is determined by identifying the class with the maximum scaled similarity score \(s_i\):


\begin{equation}
\hat{y} = \arg\max_{i} \; s_i,
\label{eq:predicted_class_clip}
\end{equation}

Finally, the accuracy is calculated as the proportion of correctly classified samples:

\begin{equation}
\text{Accuracy} = \frac{1}{M} \sum_{n=1}^{M} \mathbf{1}_{\hat{y}_n = y_n},
\label{eq:accuracy_clip}
\end{equation}

where \(M\) is the total number of samples, and \(\mathbf{1}_{\hat{y}_n = y_n}\) is an indicator function that equals 1 when the predicted class \(\hat{y}_n\) matches the true class \(y_n\).

\subsection{Sparse Low-rank Adaptation (SoRA)}~\label{sec:sora}

\vspace{-1.0em}

Sparse Low-rank Adaptation (SoRA)~\cite{ding2023sora} extends LoRA by introducing a structured pruning mechanism to further reduce the number of parameters while preserving model performance. SoRA leverages Singular Value Decomposition (SVD) to decompose the LoRA weight matrices into singular vectors, enabling the adaptive pruning of less important components during training.

SoRA begins with a higher rank for the LoRA layers, which is progressively reduced as the model identifies components that are less critical to the task. This is achieved by applying sparsity-inducing regularization to the singular values, setting unimportant components to zero and subsequently pruning them. This dynamic pruning process allows SoRA to maintain efficiency without compromising accuracy.

In our work, we adopt SoRA’s dynamic pruning approach for the image encoder of the CLIP model within the FL framework. By tailoring the pruning process to each client’s local data distribution, SoRA enables personalized model adaptation while significantly reducing communication costs. This selective pruning addresses the challenge of non-IID data in FL, allowing clients to adapt the model to their unique data while sharing only a smaller, more efficient set of parameters with the server. This approach ensures both local personalization and global generalization, improving scalability and performance in FL settings.




\subsection{FedDAT (Federated Dual-Adapter Teacher)}~\label{sec:feddat}

\vspace{-1.0em}

FedDAT~\cite{chen2024feddat} introduces a dual-adapter mechanism in Federated Learning (FL) to tackle the challenge of data heterogeneity across clients. In this approach, each client utilizes two adapters: a global adapter, shared across all clients and kept frozen during local training, and a local adapter, which is fine-tuned on the client’s specific data. This dual-adapter design enables the model to retain both client-specific knowledge for personalization and client-agnostic knowledge for generalization.

A key innovation of FedDAT is the use of Mutual Knowledge Distillation (MKD), where the local adapter learns from the frozen global adapter, and vice versa. This bidirectional knowledge transfer mitigates overfitting to local data while ensuring that the global adapter incorporates relevant client-specific information during aggregation, improving the model's ability to handle diverse data distributions.

Our method draws inspiration from FedDAT by incorporating a dual LoRA adapter setup: a local LoRA adapter tailored to each client’s data and a global LoRA adapter shared and aggregated across clients. The local LoRA is pruned based on the importance of its components for the client’s data, ensuring efficiency and adaptability to non-IID data distributions. Meanwhile, the global LoRA enables efficient knowledge sharing and collaboration among clients, analogous to the dual-adapter design in FedDAT.

This two-adapter setup is vital for handling non-IID data in FL. The local LoRA specializes in addressing client-specific data variability, enabling personalized model adaptation. At the same time, the global LoRA ensures consistency and alignment across clients by aggregating shared knowledge. This synergy between local adaptation and global collaboration enhances the model’s ability to generalize to new data distributions while maintaining computational efficiency and robustness against data heterogeneity.

\vspace{-0.5em}

\section{Methodology}~\label{sec:methodology}

\vspace{-2.0em}

\begin{figure}[ht]
\begin{center}


\centerline{\includegraphics[width=0.5\textwidth]{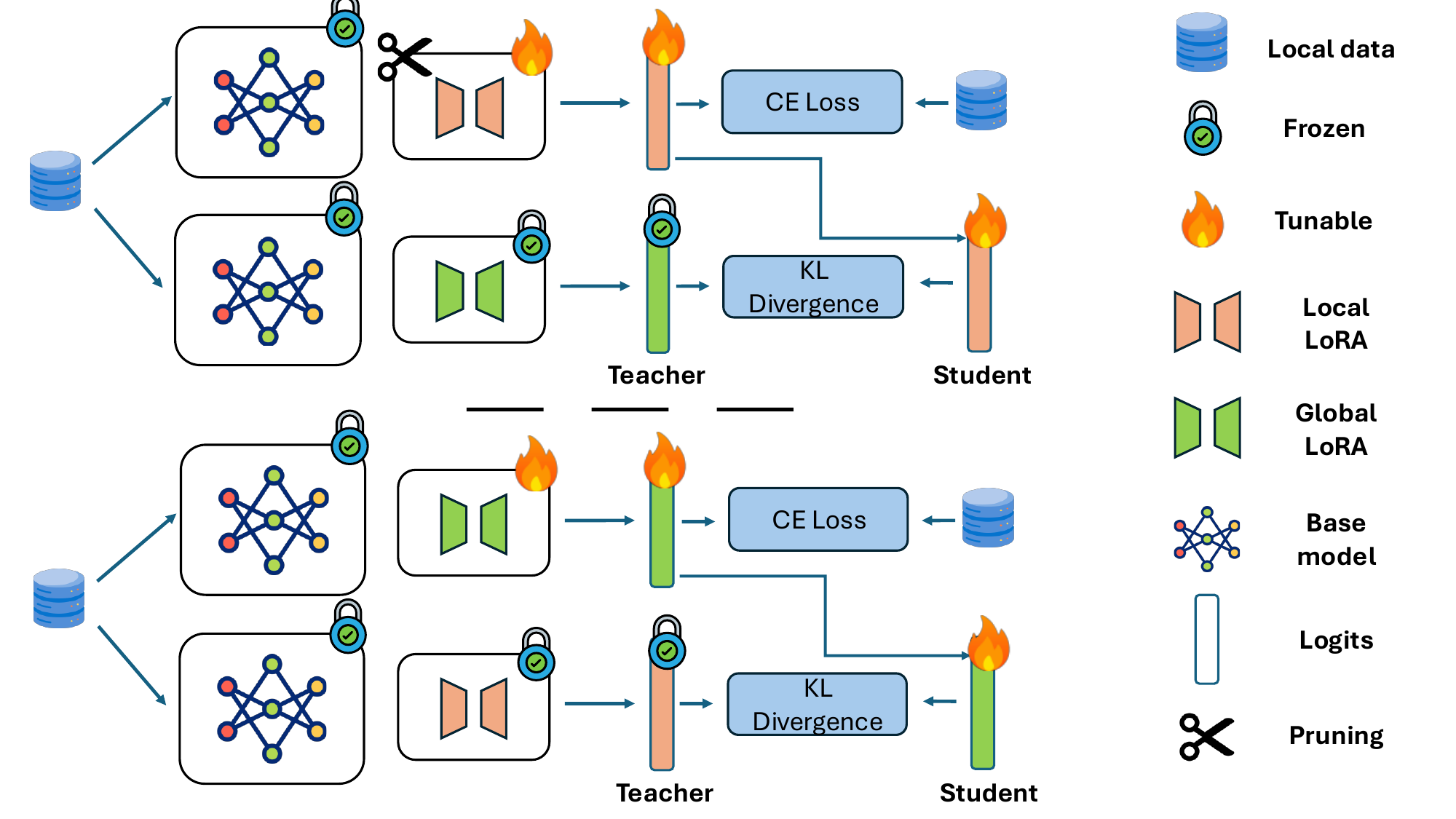}}

\caption{The overall training scheme within each client. Each client has a CLIP model with two separate LoRA adapters: a larger local LoRA (orange) and a smaller global LoRA (green). During training, the local LoRA is pruned to reduce communication overhead, and the local adapter is updated using cross-entropy (CE) loss while knowledge distillation (Kullback-Leibler (KL) divergence) is performed from the global adapter (Teacher) to the local adapter (Student). The global LoRA, which is shared across clients, is also updated using CE loss and KL divergence from the local adapter. This bi-directional distillation ensures that the global adapter retains generalizable knowledge while the local adapter remains personalized to the client's data. The pruned local adapter and the smaller global adapter are aggregated at the server for global updates. The figure is best viewed in color.
}
\vspace{-2.0em}

\label{fig:method}
\end{center}
\end{figure}

\vspace{-1.0em}

\subsection{Problem statement}

Our objective is to minimize the following global loss function across \(N\) clients:

\[
\min_{\{\theta_i\}} \sum_{i=1}^{N} \frac{|D_i|}{|D|} F_i(\Delta \cup \theta^i),
\]

where:
\begin{itemize}
    \item \(F_i(\theta^i)\) is the local loss for client \(i\), defined by the client's data distribution.
    \item \(|D_i|\) represents the size of client \(i\)'s dataset, and \(|D|\) is the total size of all clients' datasets.
    \item \(\theta^i\) are the adapter parameters for client \(i\), which include both the local and global LoRA parameters.
    \item $\Delta$ is the frozen base model.
    \item \(N\) is the number of participating clients.
    
\end{itemize}

Each client \(k\) receives the global LoRA adapter from the server, and uses it in combination with its local LoRA adapter to update its local model. The client-specific local loss function \(L_{\theta_i}\) is defined as:

\[
L_k(\theta^i) = \frac{1}{D_i} \sum_{(x, y) \in D_i} \mathcal{L}(y_i, f_{\Delta \cup \theta^i}(x_i)),
\]

where:
\begin{itemize}
    \item \(y_i\) is the ground truth label of the input data \(x_i\).
    \item \(\mathcal{L}\) is the Cross-Entropy loss function used for classification.
\end{itemize}

\subsection{Federated Learning Setup}

In our FL setup, our goal is to fine tune a base model for each client using both local and global LoRA adapters, while minimizing communication overhead and maximizing performance across heterogeneous client data. The goal of our method is to address the personalization challenges in pFL by leveraging two distinct LoRA adapters for each client, ensuring personalized model adaptation while sharing global knowledge efficiently.

As mentioned, the two-adapter setup idea is inspired by FedDAT~\cite{chen2024feddat}; however, our method introduces significant modifications, including separate local and global LoRA adapters, as well as structured pruning techniques. Figure~\ref{fig:method} depicts an overview of our method.


The \textbf{local adapter} ($\theta_l^i$) remains private to each client, enabling personalized training, while the \textbf{global adapter} ($\theta_g^i$) is periodically updated at the server using aggregated knowledge from all clients.

\subsection{Training Dynamics and Loss Functions}~\label{sec:training_procedure}

\vspace{-1.0em}

\subsubsection{Local Adapter Training}
At each client, the local LoRA adapter $\theta_l^i$ is updated based on two key components:
\begin{itemize}
    \item \textbf{Cross-Entropy (CE) Loss}: This loss ensures that the local adapter aligns with the ground truth labels in the client’s local dataset. 
    \[
    \mathcal{L}_{\text{CE}}^{\text{local}} = \text{CE}(\text{logits}_{\text{local}}, \text{labels})
    \]
    
    \item \textbf{Knowledge Distillation (KD) from Global Adapter}: In addition to the CE loss, the local adapter is trained using knowledge distillation from the frozen global adapter. This distillation enforces consistency between the local and global models by aligning their output logits through the use of Kullback-Leibler divergence ($D_{KL}$).
    \[
    \mathcal{L}_{\text{KD}}^{\text{local}} = D_{KL}(\text{logits}_{\text{local}} \parallel \text{logits}_{\text{global}})
    \]
\end{itemize}

\subsubsection{Global Adapter Training}
During each communication round, the global LoRA adapter $\theta_g^i$ is also updated at the client level, with the reverse mechanism:
\begin{itemize}
    \item \textbf{Cross-Entropy (CE) Loss}: The global adapter is optimized using cross-entropy loss on the local client’s data, allowing it to learn from the client’s task.
    \[
    \mathcal{L}_{\text{CE}}^{\text{global}} = \text{CE}(\text{logits}_{\text{global}}, \text{labels})
    \]
    
    \item \textbf{Knowledge Distillation (KD) from Local Adapter}: The global adapter is updated through knowledge distillation from the local adapter’s logits, which helps it adapt based on the client’s task.
    \[
    \mathcal{L}_{\text{KD}}^{\text{global}} = D_{KL}(\text{logits}_{\text{global}} \parallel \text{logits}_{\text{local}})
    \]
\end{itemize}

This bi-directional knowledge distillation ensures that both the local and global adapters benefit from each other’s learning without being directly combined into a single module, maintaining their independence.

\subsection{Adapter Freezing Mechanism}

To facilitate efficient learning, when updating the local LoRA adapter, the global adapter is frozen, and vice versa. This ensures that the adapters do not interfere with each other’s learning, allowing the local adapter to specialize in client-specific tasks, while the global adapter captures general trends across clients. By combining knowledge distillation, cross-entropy loss, and pruning, our method balances the benefits of both personalization, and generalization, while minimizing communication overhead in federated learning.

\subsection{Communication and Aggregation}

Once the local training phase is completed, the global LoRA adapter is communicated to the central server. At the server, all global adapters from the clients are aggregated using a weighted averaging scheme, similar to \textbf{Federated Averaging} (FedAvg)~\cite{mcmahan2017fedavg}, to update the global model. This ensures that global knowledge is retained and updated in a federated manner, while the local adapter remains personalized for each client.

\begin{table*}[ht]
\caption{The test accuracy (\%) of the image classification tasks for practical non-IID with $N=10$ and $\beta = 0.1$. \textbf{Bold} is the best performance in one specific scenario, while \underline{underline} is the second best.}
\label{tab:acc_beta01}

\vspace{-1.0em}

\begin{center}
\begin{tabular}{c|c|ccccc|c}
\toprule
\multirow{1}{*}{Adapter placement} & \multirow{1}{*}{Method} & \multirow{1}{*} {Pets $\uparrow$} & \multirow{1}{*} {Flowers $\uparrow$} & \multirow{1}{*} {Aircraft $\uparrow$} & \multirow{1}{*} {DTD $\uparrow$} & \multirow{1}{*} {Average$\uparrow$} & \multicolumn{1}{|c} {\# Params (M)$\downarrow$} \\ 
\midrule
\multirow{5}{*}{Text encoder} & Local   & 78.19$\scriptstyle{\pm 0.43}$ & 58.83$\scriptstyle{\pm 1.59}$ & 25.41$\scriptstyle{\pm 0.21}$ & 43.99$\scriptstyle{\pm 0.81}$ & 51.60$\scriptstyle{\pm 0.76}$ & - \\

& FedCLIP~\cite{lu2023fedclip}          & 81.42$\scriptstyle{\pm 0.16}$ & 80.26$\scriptstyle{\pm 0.65}$ & 31.03$\scriptstyle{\pm 0.04}$ & 49.90$\scriptstyle{\pm 0.27}$ & 60.65$\scriptstyle{\pm 0.28}$  & 0.525 \\

& FLoRA~\cite{nguyen2024flora}          & 89.06$\scriptstyle{\pm 0.24}$ & 92.76$\scriptstyle{\pm 0.36}$ & 40.83$\scriptstyle{\pm 1.14}$ & 62.22$\scriptstyle{\pm 1.04}$ & 71.22$\scriptstyle{\pm 0.69}$ & 0.223 \\

& FedDAT~\cite{chen2024feddat}          & \underline{89.51$\scriptstyle{\pm 0.62}$} & \textbf{95.57$\scriptstyle{\pm 0.07}$} & \underline{52.65$\scriptstyle{\pm 1.94}$} & \underline{78.49$\scriptstyle{\pm 0.24}$} & \underline{79.05$\scriptstyle{\pm 0.72}$} & 0.223 \\

\rowcolor{blue!10} & \proj              & \textbf{91.90$\scriptstyle{\pm 0.42}$} & \underline{95.26$\scriptstyle{\pm 0.19}$} &  \textbf{57.12$\scriptstyle{\pm 1.06}$} & \textbf{79.84$\scriptstyle{\pm 0.57}$} & \textbf{81.03$\scriptstyle{\pm 0.56}$} & 0.223 \\

\midrule

\multirow{5}{*}{Image encoder} & Local  & 82.18$\scriptstyle{\pm 0.60}$ & 55.80$\scriptstyle{\pm 0.80}$ & 25.42$\scriptstyle{\pm 0.50}$ & 26.10$\scriptstyle{\pm 0.08}$ & 47.37$\scriptstyle{\pm 0.50}$ & - \\

& FedCLIP~\cite{lu2023fedclip}          & 88.93$\scriptstyle{\pm 0.21}$ & 91.22$\scriptstyle{\pm 0.15}$ & 43.25$\scriptstyle{\pm 0.16}$ & 59.04$\scriptstyle{\pm 0.32}$ & 70.61$\scriptstyle{\pm 0.21}$ & 0.525 \\

& FLoRA~\cite{nguyen2024flora}          & 91.54$\scriptstyle{\pm 0.51}$ & 94.34$\scriptstyle{\pm 0.13}$ & 47.96$\scriptstyle{\pm 0.66}$ & 72.60$\scriptstyle{\pm 0.41}$ & 76.61$\scriptstyle{\pm 0.43}$ & 0.334 \\

& FedDAT~\cite{chen2024feddat}          & \underline{94.65$\scriptstyle{\pm 0.33}$} & \textbf{94.58$\scriptstyle{\pm 0.47}$} & \underline{59.12$\scriptstyle{\pm 0.74}$} & \underline{79.41$\scriptstyle{\pm 1.55}$} & \underline{81.94$\scriptstyle{\pm 0.77}$} & 0.334 \\

\rowcolor{blue!10} & \proj              & \textbf{94.67$\scriptstyle{\pm 0.16}$} & \underline{94.56$\scriptstyle{\pm 0.10}$} &  \textbf{59.14$\scriptstyle{\pm 0.24}$} & \textbf{81.23$\scriptstyle{\pm 0.29}$}  &  \textbf{82.40$\scriptstyle{\pm 0.20}$} & 0.334 \\

\bottomrule
\end{tabular}

\vspace{-1.0em}

\end{center}
\end{table*}

\begin{table*}[ht]
\caption{The test accuracy (\%) of the image classification tasks for practical non-IID with $N=10$ and $\beta = 0.01$. \textbf{Bold} is the best performance in one specific scenario, while \underline{underline} is the second best.}
\label{tab:acc_beta001}

\vspace{-1.0em}

\begin{center}
\begin{tabular}{c|c|ccccc|c}
\toprule
\multirow{1}{*}{Adapter placement} & \multirow{1}{*}{Method} & \multirow{1}{*} {Pets $\uparrow$} & \multirow{1}{*} {Flowers $\uparrow$} & \multirow{1}{*} {Aircraft $\uparrow$} & \multirow{1}{*} {DTD $\uparrow$} & \multirow{1}{*} {Average$\uparrow$} & \multicolumn{1}{|c} {\# Params (M)$\downarrow$} \\ 
\midrule
\multirow{5}{*}{Text encoder} & Local   & 79.51$\scriptstyle{\pm 0.57}$ & 53.86$\scriptstyle{\pm 0.48}$ & 8.50$\scriptstyle{\pm 0.04}$ & 34.32$\scriptstyle{\pm 2.22}$ & 44.05$\scriptstyle{\pm 0.83}$ & - \\

& FedCLIP~\cite{lu2023fedclip}          & 75.45$\scriptstyle{\pm 0.16}$ & 71.76$\scriptstyle{\pm 0.00}$ & 30.37$\scriptstyle{\pm 0.52}$ & 47.60.$\scriptstyle{\pm 0.47}$ & 56.29$\scriptstyle{\pm 0.28}$  & 0.525 \\

& FLoRA~\cite{nguyen2024flora}          & 80.81$\scriptstyle{\pm 0.24}$ & 84.44$\scriptstyle{\pm 1.09}$ & 35.10$\scriptstyle{\pm 1.30}$ & 54.75$\scriptstyle{\pm 0.93}$ & 63.77$\scriptstyle{\pm 0.89}$ & 0.223 \\

& FedDAT~\cite{chen2024feddat}          & \underline{89.37$\scriptstyle{\pm 1.51}$} & \underline{93.43$\scriptstyle{\pm 1.51}$} & \underline{50.01$\scriptstyle{\pm 3.08}$} & \underline{79.70$\scriptstyle{\pm 3.11}$} & \underline{78.13$\scriptstyle{\pm 2.30}$} & 0.223 \\

\rowcolor{blue!10} & \proj              & \textbf{90.98$\scriptstyle{\pm 0.96}$} & \textbf{95.78$\scriptstyle{\pm 0.96}$} &  \textbf{58.26$\scriptstyle{\pm 1.98}$} & \textbf{81.31$\scriptstyle{\pm 0.88}$} & \textbf{81.58$\scriptstyle{\pm 1.18}$} & 0.223 \\

\midrule

\multirow{5}{*}{Image encoder} & Local  & 80.42$\scriptstyle{\pm 0.23}$ & 57.12$\scriptstyle{\pm 0.28}$ & 12.79$\scriptstyle{\pm 0.59}$ & 24.93$\scriptstyle{\pm 0.32}$ & 43.81$\scriptstyle{\pm 0.36}$ & - \\

& FedCLIP~\cite{lu2023fedclip}          & 81.21$\scriptstyle{\pm 0.46}$ & 85.53$\scriptstyle{\pm 0.31}$ & 39.68$\scriptstyle{\pm 0.24}$ & 57.60$\scriptstyle{\pm 0.14}$ & 66.00$\scriptstyle{\pm 0.29}$ & 0.525 \\

& FLoRA~\cite{nguyen2024flora}          & 87.68$\scriptstyle{\pm 0.60}$ & 91.17$\scriptstyle{\pm 0.60}$ & 44.94$\scriptstyle{\pm 2.03}$ & 69.26$\scriptstyle{\pm 0.74}$ & 73.26$\scriptstyle{\pm 1.00}$ & 0.334 \\

& FedDAT~\cite{chen2024feddat}          & \underline{96.12$\scriptstyle{\pm 0.70}$} & \underline{97.51$\scriptstyle{\pm 0.70}$} & \underline{61.65$\scriptstyle{\pm 0.24}$} & \underline{84.11$\scriptstyle{\pm 0.89}$} & \underline{84.85$\scriptstyle{\pm 0.63}$} & 0.334 \\

\rowcolor{blue!10} & \proj              & \textbf{96.44$\scriptstyle{\pm 0.29}$} & \textbf{97.85$\scriptstyle{\pm 0.29}$} &  \textbf{67.52$\scriptstyle{\pm 1.55}$} & \textbf{86.81$\scriptstyle{\pm 1.37}$}  &  \textbf{87.15$\scriptstyle{\pm 0.88}$} & 0.334 \\

\bottomrule
\end{tabular}

\vspace{-3.0em}

\end{center}
\end{table*}

\subsection{Combined Loss Function for LoRA Adapters}


\subsubsection{Loss Function for Local Adapter}

For the local LoRA adapter $\theta_l^i$, the overall loss function combines the Cross-Entropy (CE) loss with the KD loss. Since the local adapter has more capacity than the global adapter, we introduce a hyperparameter \(\alpha\) to control the relative contribution of the KD loss, which aligns the local adapter's output with the global adapter’s predictions:

\[
\mathcal{L}^{\text{local}} = \mathcal{L}_{\text{CE}}^{\text{local}} + \alpha \cdot \mathcal{L}_{\text{KD(global)}}^{\text{local}},
\]

where:
\begin{itemize}
    \item \(\mathcal{L}_{\text{CE}}^{\text{local}}\) is the cross-entropy loss for classification using the local adapter’s logits.
    \item \(\mathcal{L}_{\text{KD(global)}}^{\text{local}} = D_{KL}(\text{logits}_{\text{local}} \parallel \text{logits}_{\text{global}})\) is the KD loss that aligns the logits of the larger local adapter with the logits of the global adapter.
\end{itemize}

\subsubsection{Loss Function for Global Adapter}

For the global LoRA adapter $\theta_g^i$, the loss function includes both CE loss and KD loss. However, since the global adapter has less capacity, no control hyper parameter is introduced for the KD loss to help stabilize performance across rounds, and the loss function is defined as:

\[
\mathcal{L}^{\text{global}} = \mathcal{L}_{\text{CE}}^{\text{global}} + \mathcal{L}_{\text{KD(local)}}^{\text{global}},
\]

where:
\begin{itemize}
    \item \(\mathcal{L}_{\text{CE}}^{\text{global}}\) is the cross-entropy loss for classification using the global adapter’s logits.
    \item \(\mathcal{L}_{\text{KD(local)}}^{\text{global}} = D_{KL}(\text{logits}_{\text{global}} \parallel \text{logits}_{\text{local}})\) is the KD loss that aligns the logits of the global adapter with the logits of the larger local adapter.
\end{itemize}

\subsection{Pruning for the Local LoRA Adapters}

We incorporate the idea of SoRA, as proposed by Ding et al.~\cite{ding2023sora}, to efficiently reduce the memory and computation costs associated with the larger local LoRA adapter. This pruning process zeroes out unimportant weights in the local adapter during training, enabling clients to maintain computational efficiency while maximizing performance. The pruning process is defined as:

\[
    \mathbf{g}_{t+1}^{i} \gets \mathcal{T}_{\xi} \left( \mathbf{g}_t^i - \mu \nabla_{\mathbf{g}^{i}} \mathcal{L}^{local} (\theta^i_t) \right),
    \]
    

    where \( \mathbf{g}^i \) represents the gate of the SoRA module at the \(i\)-th client, \( \mathcal{T} \) is the broadcasting threshold function, and \( \xi \) is a regularization parameter. The broadcasting threshold function \( \mathcal{T} \) ensures that weights below a certain importance threshold are zeroed out, enforcing sparsity in the local adapter. This approach dynamically adjusts the number of trainable parameters, retaining only the most critical components for effective learning. By leveraging this adaptive mechanism, the model optimizes memory usage and computational resources while preserving the capacity for both personalization and generalization.


In addition to reducing the size of the local LoRA, this pruning process improves convergence by mitigating overfitting on the local dataset.


\subsection{Comparison to Related Work}

Our method builds upon the core ideas of FedDAT~\cite{chen2024feddat} but introduces key distinctions:
\begin{itemize}
    \item Unlike FedDAT, where local and global LoRA adapters are combined into a single model, we maintain \textbf{separate} adapters for local and global representations, ensuring personalized local training. 
    
    \item We introduce \textbf{pruning} for the local adapter, leveraging ideas from SoRA to enhance efficiency and reduce computation.
\end{itemize}

\begin{table*}[ht]
\caption{The test accuracy distribution among individual clients with $N=10$ and $\beta = 0.1$. Standard deviation ($\times$ 100) and lowest accuracy (\%) of local performances. \textbf{Bold} is the best performance in one specific scenario, while \underline{underline} is the second best.}
\label{tab:std_acc_beta01}

\vspace{-1.0em}

\begin{center}
\begin{tabular}{c|c|cccccccc}
\toprule
\multirow{2}{*}{Adapter placement} & \multirow{2}{*}{Method} & \multicolumn{2}{c} {Pets} & \multicolumn{2}{c} {Flowers} & \multicolumn{2}{c} {Aircraft} & \multicolumn{2}{c} {DTD} \\ 
 &  & {std $\downarrow$} & {Lowest acc. $\uparrow$} & {std $\downarrow$} & {Lowest acc. $\uparrow$} & {std $\downarrow$} & {Lowest acc. $\uparrow$} & {std $\downarrow$} & {Lowest acc. $\uparrow$} \\ 
\midrule
\multirow{5}{*}{Text encoder} & Local   & 4.08 & 70.53 & 10.29 & 50.50 & \textbf{4.34} & 20.15 & 9.82 & 33.07 \\

& FedCLIP~\cite{lu2023fedclip}          & 5.89 & 73.42 & 5.31 & 69.90 & 6.22 & 20.45 & 11.18 & 30.40 \\

& FLoRA~\cite{nguyen2024flora}          & 3.92 & 80.92 & \textbf{2.58} & 86.11 & 5.46 & 30.25 & 10.28 & 44.56 \\

& FedDAT~\cite{chen2024feddat}          & \underline{3.29} & \underline{82.89} & \underline{2.59} & \textbf{90.00} & 5.77 & 42.05 & \textbf{4.86} & \underline{70.65} \\

\rowcolor{blue!10} & \proj              & \textbf{3.13} & \textbf{86.18} & 3.01 & \underline{89.13} & \underline{5.28} & \textbf{51.37} & \underline{6.45} & \textbf{71.73} \\

\midrule

\multirow{5}{*}{Image encoder} & Local  & 2.54 & 77.29 & 11.16 & 44.50 & 7.26 & 14.22 & 15.85 & 12.30 \\

& FedCLIP~\cite{lu2023fedclip}          & 4.18 & 80.92 & 4.86 & 81.94 & 5.56 & 34.88 & 9.97 & 41.89 \\

& FLoRA~\cite{nguyen2024flora}          & 4.02 & 84.05 & 3.83 & 87.50 & 6.78 & 35.89 & \underline{7.45} & 62.32 \\

& FedDAT~\cite{chen2024feddat}          & \underline{1.66} & \textbf{92.27} & \underline{2.53} & \underline{91.30} & \underline{4.85} & \underline{51.79} & 7.57 & \underline{64.70} \\

\rowcolor{blue!10} & \proj              & \textbf{1.39} & \underline{91.48} & \textbf{1.97} & \textbf{91.77} & \textbf{3.23} & \textbf{54.92} & \textbf{7.06} & \textbf{71.24} \\

\bottomrule
\end{tabular}

\vspace{-1.0em}

\end{center}
\end{table*}

\begin{table*}[ht]
\caption{The test accuracy distribution among individual clients with $N=10$ and $\beta = 0.01$. Standard deviation ($\times$ 100) and lowest accuracy (\%) of local performances. \textbf{Bold} is the best performance in one specific scenario, while \underline{underline} is the second best.}

\vspace{-1.0em}

\label{tab:std_acc_beta001}
\begin{center}
\begin{tabular}{c|c|cccccccc}
\toprule
\multirow{2}{*}{Adapter placement} & \multirow{2}{*}{Method} & \multicolumn{2}{c} {Pets} & \multicolumn{2}{c} {Flowers} & \multicolumn{2}{c} {Aircraft} & \multicolumn{2}{c} {DTD} \\ 
 &  & {{std} $\downarrow$} & {Lowest acc. $\uparrow$} & {{std} $\downarrow$} & {Lowest acc. $\uparrow$} & {{std} $\downarrow$} & {Lowest acc. $\uparrow$} & {{std} $\downarrow$} & {Lowest acc. $\uparrow$} \\ 
\midrule
\multirow{5}{*}{Text encoder} & Local   & \textbf{5.57} & 63.09 & 9.75 & 42.55 & \textbf{1.41} & 6.59 & \underline{9.85} & 18.90 \\

& FedCLIP~\cite{lu2023fedclip}          & 16.05 & 30.09 & 8.05 & 58.51 & 6.99 & 20.70 & 14.48 & 20.00 \\

& FLoRA~\cite{nguyen2024flora}          & 12.54 & 54.36 & 9.73 & 63.90 & 9.43 & 25.68 & 17.13 & 25.88 \\

& FedDAT~\cite{chen2024feddat}          & 9.65 & \underline{67.56} & \underline{3.48} & \underline{89.04} & \underline{6.75} & \underline{43.07} & 12.46 & \underline{59.86} \\

\rowcolor{blue!10} & \proj              & \underline{6.43} & \textbf{77.66} & \textbf{1.89} & \textbf{93.17} & 7.64 & \textbf{49.77} & \textbf{5.18} & \textbf{69.86} \\

\midrule

\multirow{5}{*}{Image encoder} & Local  & \textbf{2.29} & \textbf{77.25} & 8.93 & 45.74 & \textbf{4.87} & 5.12 & 15.79 & 12.80 \\

& FedCLIP~\cite{lu2023fedclip}          & \underline{10.85} & \underline{63.10} & 5.68 & 72.03 & 9.40 & 24.34 & 19.09 & 28.23 \\

& FLoRA~\cite{nguyen2024flora}          & 14.57 & 45.63 & 3.50 & 85.06 & 9.52 & 32.58 & 15.32 & 36.47 \\

& FedDAT~\cite{chen2024feddat}          & 13.54 & 52.72 & \underline{1.81} & \underline{94.52} & 9.93 & \underline{50.18} & \underline{12.46} & \underline{62.35} \\

\rowcolor{blue!10} & \proj              & 16.13 & 44.66 & \textbf{1.62} & \textbf{94.68} & \underline{8.96} & \textbf{54.68} & \textbf{10.47} & \textbf{63.52} \\

\bottomrule
\end{tabular}

\vspace{-2.0em}

\end{center}
\end{table*}

\vspace{-1.0em}

\section{Experimental Setup}~\label{sec:experiment}

\vspace{-1.0em}

In our experiments, we evaluate the \proj framework on various datasets, and compare its performance against baseline FL approaches, including:

\begin{itemize}

    \item \textbf{Local Training Only (Local)}: In this setup, each client trains its model locally without exchanging LoRA layers with other clients or the server. Similar to ~\cite{nguyen2024flora}, we set LoRA adapter with the rank of 2. This serves as a benchmark to assess the benefit of FL versus isolated local training.

    \item \textbf{FedCLIP}~\cite{lu2023fedclip}: This approach adds an adapter (consisting of two fully connected layers) at the end of either the text, or image encoder of the CLIP model and fine-tunes it.

    \item \textbf{Full LoRA Application (FLoRA)}~\cite{nguyen2024flora}: In this method, LoRA is applied to all linear modules of the CLIP encoder (either text or image). It fine-tunes the entire set of linear layers in the encoder with a LoRA rank of 2.

    \item \textbf{FedDAT}~\cite{chen2024feddat}:  This method employs dual LoRA adapters with Mutual Knowledge Distillation. The adapters are applied to all linear modules of the CLIP encoder, and both LoRA adapters are initialized with the same rank of 2.

    \item \textbf{\proj (Our Method)}: For a fair comparison with FedDAT, which uses dual adapters, \proj initializes the local LoRA adapter with a rank of 4, and the global adapter with a rank of 2. The local adapter is pruned during training to reduce communication overhead while preserving performance.
\end{itemize}
    
\noindent For all methods, the CLIP base model remains frozen during the federated training process. The experiments are conducted separately for both the text and image encoders, following prior work on CLIP adapters~\cite{gao2021clipadapter}.

\vspace{-0.5em}

\subsection{Datasets and Metrics} 
We use publicly available datasets for vision and language tasks to evaluate our framework: Oxford-IIIT Pet~(Pets)~\cite{parkhi2012pets}, Oxford 102 Flower~(Flowers)~\cite{nilsback2008flowers}, FGVC-Aircraft (Aircraft)~\cite{maji2013aircraft}, and Describable Textures Dataset~(DTD)~\cite{cimpoi14dtd}. The performance metrics used to evaluate the methods include: The average classification performance on an individual testing set, and the total amount of data communicated between the clients and the server until reaching a target accuracy.

\subsubsection{Data Partitioning for FL}

To simulate realistic FL scenarios, we use non-IID data partitioning with a Dirichlet distribution, which allows us to control the degree of data heterogeneity across clients~\cite{lin2020feddf, li2021moon}. A smaller 
$\beta$ value results in more heterogeneous data. We consider two settings: moderate heterogeneity ($\beta=0.1$), where client data distributions vary but are not entirely disjoint, and high heterogeneity ($\beta=0.01$), where data skewness is extreme, with many classes absent from certain clients, reflecting real-world scenarios.



To simulate the pFL setting, all datasets are split randomly with 75\% and 25\% for training and testing, respectively. We evaluate the performance of each pFL algorithm using the average local accuracy as the metric. To ensure robustness, we report results for 100 rounds with mean ± standard deviation across three random seeds (0, 1, 42).

\subsubsection{Experimental Procedure}

The federated training process is executed for 100 rounds, with each of the 10 clients performing one local epoch per round. Similar to ~\cite{lu2023fedclip}, the batch size is set to 32. We use AdamW optimizer. All experiments are conducted on a single GPU A100 with 40 GB of memory. The pre-trained CLIP model utilized is based on ViT-B/32 \cite{dosovitskiy2021vit} as the base image encoder.

To identify the optimal learning rates for the LoRA adapters and other hyper parameters, we perform a grid search. The search space for the learning rate of the adapter in FedCLIP includes: $\{5\times10^{-3}, 1\times10^{-3}, 5\times10^{-4}, 1\times10^{-4}, 5\times10^{-5}, 1\times10^{-5}\}$. The search space for the learning rate of all LoRA adapters includes: $\{5\times10^{-4}, 1\times10^{-4}, 5\times10^{-5}, 1\times10^{-5}, 5\times10^{-6}, 1\times10^{-6}, 5\times10^{-7}, 1\times10^{-7}\}$. For the control parameter of $\alpha$, the search space includes $\{100, 10, 1, 0.1, 0.01, 0.001\}$. We evaluate the performance of each combination on a testing set to determine the best settings for our experiments. We set $\xi=5\times10^{-5}$ as the threshold for gate pruning in SoRA.

\section{Results and Analysis}~\label{sec:results}

\vspace{-1.0em}


\noindent \textbf{Accuracy Comparison.} Tables~\ref{tab:acc_beta01} and~\ref{tab:acc_beta001} summarize the test accuracy for text and image encoders under varying levels of data heterogeneity ($\beta = 0.1$ and $\beta = 0.01$). Across both settings, \proj demonstrates the highest average accuracy across most datasets, outperforming all baselines. For instance, with the text encoder and $\beta = 0.1$ on DTD, \proj achieves an average accuracy of 79.84\%, compared to 79.49\% for FedDAT and 72.60\% for FLoRA. Although FedDLP is narrowly outperformed by FedDAT on the Flowers dataset with $\beta = 0.1$ (95.78\% vs. 95.26\%), the margin is minimal, highlighting the competitive performance of \proj. Similar trends are observed for $\beta = 0.01$, where \proj consistently achieves superior performance, affirming its robustness across heterogeneous data distributions.

Interestingly, both FedDLP and, in some cases, FedDAT show improved accuracy as the data heterogeneity increases (i.e., when $\beta$ decreases from 0.1 to 0.01). For instance, on the Aircraft dataset, FedDLP achieves a higher accuracy with $\beta = 0.01$, suggesting that the pruning and dual-adapter mechanisms in FedDLP effectively mitigate the challenges posed by extreme heterogeneity. This improvement could be attributed to the ability of the dual-adapter architecture to balance local personalization and global generalization, enabling the model to better capture the unique patterns in highly skewed client data. Similarly, FedDAT benefits from its mutual knowledge distillation strategy, which aligns local and global representations, enhancing performance under increased heterogeneity.


\begin{table}[ht]
\caption{Comparing communication costs in MB of different methods to reach desired target accuracy. Costs relative to
FedDAT are in parentheses. Adapters are applied to the text encoder. \textbf{Bold} numbers highlight lowest overhead. }
\label{tab:cost_text}

\vspace{-1.0em}

\begin{center}
\begin{tabular}{c|c|c|ccc}
\toprule
\multirow{1}{*}{Dataset} & \multirow{1}{*}{$\beta$} & \multirow{1}{*}{Target Acc.} & \multirow{1}{*} {FedDAT} & \multirow{1}{*} {FedDLP} \\ 
\midrule

    \multirow{2}{*}{Pets} & \multirow{1}{*} {0.1} & \multirow{1}{*} {$88\%$} & 34.08 & \textbf{13.63 (0.40$\times$)} \\

    & \multirow{1}{*} {0.01} & \multirow{1}{*} {$85\%$} & {20.48} & \textbf{17.04 (0.83$\times$)} \\

    \midrule

    \multirow{2}{*}{Flowers} & \multirow{1}{*} {0.1} & \multirow{1}{*} {$91\%$} & 15.34 & \textbf{11.93 (0.78$\times$)} \\

    & \multirow{1}{*} {0.01} & \multirow{1}{*} {$90\%$} & 20.45 & \textbf{11.93 (0.58$\times$)} \\

    \midrule

    \multirow{2}{*}{Aircraft} & \multirow{1}{*} {0.1} & \multirow{1}{*} {$52\%$} & 97.13 & \textbf{27.26 (0.46$\times$)} \\
  
     & \multirow{1}{*} {0.01} & \multirow{1}{*} {$50\%$} & 132.91 & \textbf{34.08 (0.26$\times$)} \\

    \midrule

    \multirow{2}{*}{DTD} & \multirow{1}{*} {0.1} & \multirow{1}{*} {$75\%$} & 47.71 & \textbf{27.26 (0.57$\times$)} \\
    & \multirow{1}{*} {0.01} & \multirow{1}{*} {$75\%$} & 28.97 & \textbf{20.45 (0.71$\times$)} \\

\bottomrule

\end{tabular}

\vspace{-2.0em}

\end{center}
\end{table}

\begin{table}[ht]
\caption{Comparing communication costs in MB of different methods to reach desired target accuracy. Costs relative to FLoRA are in parentheses. Adapters are applied to the image encoder. \textbf{Bold} numbers highlight lowest overhead.}
\label{tab:cost_image}

\vspace{-1.0em}

\begin{center}
\begin{tabular}{c|c|c|ccc}
\toprule
\multirow{2}{*}{Dataset} & \multirow{2}{*}{$\beta$} & \multirow{1}{*}{Target} & \multirow{2}{*} {FLoRA} & \multirow{2}{*} {FedDAT} & \multirow{2}{*} {FedDLP} \\ 

& & \multirow{1}{*}{Acc.} & & & \\

\midrule

    \multirow{2}{*}{Pets} & \multirow{1}{*} {0.1} & \multirow{1}{*} {$88\%$} & 28.05 & 15.30 (0.54$\times$) & \textbf{10.20 (0.37$\times$)} \\
 
    & \multirow{1}{*} {0.01} & \multirow{1}{*} {$88\%$} & 25.50 & 22.95 ($0.90\times$) & \textbf{10.20 (0.40$\times$)} \\

    \midrule

    \multirow{2}{*}{Flowers} & \multirow{1}{*} {0.1} & \multirow{1}{*} {$90\%$} & 48.45 & 25.50 (0.53$\times$) & \textbf{15.30 (0.32$\times$)} \\
    
    & \multirow{1}{*} {0.01} & \multirow{1}{*} {$90\%$} & 53.55 & 20.40 (0.38$\times$) & \textbf{17.85 (0.33$\times$)} \\

    \midrule

    \multirow{2}{*}{Aircraft} & \multirow{1}{*} {0.1} & \multirow{1}{*} {$47\%$} & 109.65 & 40.80 (0.37$\times$) & \textbf{20.40 (0.19$\times$)} \\
    
    & \multirow{1}{*} {0.01} & \multirow{1}{*} {$44\%$} & 48.45 & 30.60 (0.63$\times$) & \textbf{7.65 (0.16$\times$)} \\

    \midrule

    \multirow{2}{*}{DTD} & \multirow{1}{*} {0.1} & \multirow{1}{*} {$70\%$} & 68.85 & 56.10 (0.81$\times$) & \textbf{25.50 (0.37$\times$)} \\
    & \multirow{1}{*} {0.01} & \multirow{1}{*} {$70\%$} & 40.80 & 35.70 (0.87$\times$) & \textbf{10.20 (0.25$\times$)} \\

\bottomrule
\end{tabular}

\vspace{-3.0em}

\end{center}
\end{table}

\noindent \textbf{Client Variability.} Tables~\ref{tab:std_acc_beta01} and~\ref{tab:std_acc_beta001} provide insights into the accuracy distribution across individual clients by reporting the standard deviation and lowest accuracy. In general, \proj achieves lower standard deviations and higher lowest accuracies compared to other methods, demonstrating its ability to balance performance across heterogeneous clients. For instance, with the text encoder and $\beta = 0.1$, \proj achieves a standard deviation of 3.13 on the Pets dataset and a lowest accuracy of 86.18\%, highlighting its effectiveness in reducing variability while maintaining strong performance. However, in some cases, local training methods exhibit slightly lower standard deviations but at the cost of significantly lower average and lowest accuracies, making them less suitable for federated settings where both consistency and high performance are essential.


\noindent \textbf{Computation and Communication Costs.} Since local training, FedCLIP, and occasionally FLoRA fail to reach target accuracies in various settings, we choose FedDAT as the baseline when applying LoRA to text encoder, and FLoRA as the baseline when applying LoRA to image encoder. Tables~\ref{tab:cost_text} and~\ref{tab:cost_image} compare the computation and communication costs for reaching the target accuracy. Across all datasets, \proj achieves the lowest overhead in both computation and communication. For example, when targeting 90\% accuracy on the Flowers dataset with $\beta = 0.1$, \proj requires only 15.30 MB, compared to 25.50 MB required by FedDAT and 48.45 MB required by FLoRA with adapters being applied to image encoder. Similarly, \proj achieves substantial reductions in communication costs, requiring only 0.78$\times$ of the costs compared to FedDAT for text encoders, and 0.32$\times$ of the costs compared to FLoRA for image encoders.

\noindent \textbf{Summary of Results.}
\proj’s superior performance stems from the synergy between the dual-adapter design and pruning mechanism. The local adapter enables personalized updates, while the global adapter facilitates generalization. Pruning reduces redundancy and overfitting, focusing on critical parameters. This balance improves accuracy on challenging datasets like Aircraft and makes fine-tuning efficient by avoiding noise.


\section{Ablations}~\label{sec:ablation}

\vspace{-1.0em}

To evaluate the effectiveness of our framework, we conducted an ablation study by combining the local and global adapters into a single unified model, similar to the approach used in FedDAT. This combined setup was tested on the same datasets and configurations to understand the impact of separating the adapters.

The results, shown in Tables~\ref{tab:ablation_0.1} and~\ref{tab:ablation_0.01}, highlight the advantages of maintaining separate local and global adapters. Specifically, our framework consistently outperforms the combined approach, particularly in scenarios with high data heterogeneity (\(\beta=0.01\)). For instance, in Table~\ref{tab:ablation_0.01}, \proj achieves 97.51\% accuracy on the Flowers dataset and 86.48\% on DTD using the image encoder, compared to 93.80\% and 75.51\% for the combined setup. These improvements demonstrate the critical role of independent tuning for local and global adapters in addressing the challenges of non-IID data.

Separating the adapters enables the model to balance personalized learning (local adapter) and global knowledge sharing (global adapter), which is essential for effectively handling heterogeneous client data. In contrast, combining the adapters risks diluting these effects, particularly under extreme heterogeneity, leading to reduced accuracy and less efficient generalization.






\begin{table}[ht]
\caption{The test accuracy (\%) of the image classification tasks for practical non-IID with $N=10$ and $\beta = 0.1$.}
\label{tab:ablation_0.1}

\vspace{-1.0em}

\begin{center}
\begin{tabular}{c|c|cccc}
\toprule
\multirow{1}{*}{Adapter} & \multirow{2}{*}{Method} & \multirow{2}{*} {Pets} & \multirow{2}{*} {Flowers} & \multirow{2}{*} {Aircraft} & \multirow{2}{*} {DTD} \\ 

\multirow{1}{*}{Placement} &  &  &  &  &  \\ 
\midrule
\multirow{1}{*}{Text} & Combined & 89.57 & 94.44 & 53.21 & 71.21 \\

\multirow{1}{*}{Encoder} & \proj & \textbf{91.63} & \textbf{95.07} &  \textbf{55.92} & \textbf{79.27} \\

\midrule

\multirow{1}{*}{Image} & Combined & 93.59 & 93.71 & 54.57 & 77.01 \\

\multirow{1}{*}{Encoder} & \proj  & \textbf{94.51} & \textbf{94.68} &  \textbf{59.20} & \textbf{81.32} \\

\bottomrule
\end{tabular}

\vspace{-2.0em}

\end{center}
\end{table}

\begin{table}[ht]
\caption{The test accuracy (\%) of the image classification tasks for practical non-IID with $N=10$ and $\beta = 0.01$.}
\label{tab:ablation_0.01}

\vspace{-1.0em}

\begin{center}
\begin{tabular}{c|c|cccc}
\toprule
\multirow{1}{*}{Adapter} & \multirow{2}{*}{Method} & \multirow{2}{*} {Pets} & \multirow{2}{*} {Flowers} & \multirow{2}{*} {Aircraft} & \multirow{2}{*} {DTD} \\ 

\multirow{1}{*}{Placement} &  &  &  &  &  \\ 

\midrule

\multirow{1}{*}{Text} & Combined & 81.59 & 91.85 & 57.08 & 71.40 \\

\multirow{1}{*}{Encoder} & \proj & \textbf{90.06} & \textbf{96.24} &  \textbf{58.52} & \textbf{82.09} \\

\midrule

\multirow{1}{*}{Image} & Combined & 89.03 & 93.80 & 50.65 & 75.51 \\

\multirow{1}{*}{Encoder} & \proj & \textbf{95.43} & \textbf{97.51} &  \textbf{68.26} & \textbf{86.48} \\

\bottomrule
\end{tabular}

\vspace{-2.0em}

\end{center}
\end{table}

\section{Conclusion}~\label{sec:conclusion}

\vspace{-1.0em}

In this paper, we introduced \proj, a novel FL framework that enhances both efficiency and accuracy in personalized FL. By combining a dual-adapter approach with a pruning mechanism, \proj balances local model specialization and global knowledge sharing while significantly reducing communication overhead and improving scalability, making it practical for real-world FL with heterogeneous data.

Extensive experiments demonstrate that \proj consistently outperforms state-of-the-art methods while minimizing computational and communication costs, addressing key challenges in scaling FL systems. Future work will explore extending \proj to multi-modal tasks, asynchronous settings, and advanced pruning techniques to further optimize the trade-off between personalization and generalization.

\bibliographystyle{IEEEtranS}
\bibliography{refs}

\end{document}